# Speeding up the construction of slow adaptive walks

Susan Khor

**Abstract**

An algorithm (*bliss*) is proposed to speed up the construction of slow adaptive walks. Slow adaptive walks are adaptive walks biased towards closer points or smaller move steps. They were previously introduced to explore a search space, e.g. to detect potential local optima or to assess the ruggedness of a fitness landscape. To avoid the quadratic cost of computing Hamming distance (HD) for all-pairs of strings in a set in order to find the set of closest strings for each string, strings are sorted and clustered by *bliss* such that similar strings are more likely to get paired off for HD computation. To efficiently arrange the strings by similarity, *bliss* employs the idea of shared non-overlapping position specific subsequences between strings which is inspired by an alignment-free protein sequence comparison algorithm. Tests are performed to evaluate the quality of *b-walks*, i.e. slow adaptive walks constructed from the output of *bliss*, on enumerated search spaces. Finally, *b-walks* are applied to explore larger search spaces with the help of Wang-Landau sampling.

## 1. The *bliss* algorithm

To reduce the computation cost of constructing slow adaptive walks [1], we propose the *bliss* algorithm outlined in Fig. 1. The idea underlying *bliss* is to compute Hamming distance (HD) only for those pairs of strings which are highly similar so that we reduce the number of pairs evaluated. *bliss* arranges strings in the input set into clusters of strings having common non-overlapping position specific subsequences. This technique is inspired by an alignment-free protein sequence comparison algorithm [2]. However *bliss* avoids the costly join phase of the *afree* algorithm in ref. [2] by using subsequence exclusion. This modification is explained further on in this section (Fig. 8).

To describe *bliss* in detail, we use the problem of finding neighbor strings for each string in S = {0101, 0011, 1010, 1101, 1011, 1000} where N = 4. Strings in S are indexed from 0 to 5, i.e. the sid of '0101' is 0 and the sid of '1000' is 5 or $101_2$. For this demonstration, the sid of a string is also its fitness, i.e. F('0011') = 1 and F('1011') is 4. Each string s in S is divided into two since w = N/2, and hence wid values are either 0 or 1 to represent the two windows or subsequences. The pattern p used to extract subsequences in Fig. 2a is 1100, and in Fig. 2b is 0110. Use of other patterns is discussed with ref. to Figs. 4, 5 & 6.

The table at the top of Fig. 2a gives the hamming distance (HD) between all pairs of strings in S and the ideal neighbor set for each s, i.e. strings in S which are closest (minimum HD) to s. E.g. the nearest string to string 3 are strings 4 and 5. We would like *bliss* to find these neighbor sets, through the V sets, but avoid having to compute HD for all pairs of strings in S. *bliss* helps in this regard by suggesting pairs of similar strings. In what follows, we use the example in Fig. 2a to describe the *bliss* algorithm in detail, and return to Fig. 2b later on in this section.

---

INPUTS:
S, a set of |S| {0, 1} strings of length N. Each string is uniquely identified by a sid. Each string s in S is mapped to a real number by a fitness function F. w, the length of a subsequence, such that w < N and w divides N. Each subsequence is uniquely identified by a wid. p, a pattern of length N assigning each character in a string to a subsequence.

OUTPUT:
For each string s in S, a set V = {v | HD(s, v) is b-minimal and F(v) > F(s)}. $V_{p,w}(s)$ is the slow adaptive walk neighborhood for s given pattern p and length w. HD(s, v) is b-minimal when for all strings that *bliss* compares s with, v is one of the strings which has minimum Hamming distance. The fittest strings in S have empty V sets.

PROCEDURE:
For each string s in S, produce N/w subsequences of length N-w by excluding one subsequence of length w each time. Let y denote the N-w subsequence, and x the excluded w subsequence. To each y subsequence, append the wid of the excluded subsequence x, and the sid y and x originated from. This step results in a table of |S| N/w rows of length N-w + $\log_2 \lceil N/w \rceil$ + $\log_2 \lceil |S| \rceil$ bits.

Sort the table with LSD radix sort on (y) subsequence and wid. The sort reveals clusters of strings with identical (y) subsequence and wid.

Sequentially from one end to the other of the table, construct V sets from each cluster. A cluster of one is merged with the succeeding cluster if one exists. So are un-neighbored sequences in the current cluster, i.e. the fittest sequence(s) in a cluster.

---

**Fig. 1** The *bliss* algorithm

In the Fig. 2a example, since p = 1100, the last two characters of a string are excluded, leaving the rest of the characters included in the first window or wid = 0 for each string s in S. For the second window, the first two characters are excluded and the other characters included. E.g.: s = '1101', the subsequence for wid = 0 is '11', and for wid = 1 is '01'. Each included subsequence along with their sid and wid in binary form are placed as a record in a table. E.g.: the record 00**01**1011 in the pre-sorted table in Fig. 2a is (starting from the right), the subsequence '11' for wid = 0 and sid = $011_2$. Use of the excluded subsequence may seem redundant here where there are only two windows. However, its utility comes into play when there are more than two windows or subsequences per string, as explained later in this section.

The table of bit records is sorted using LSD radix sort on a key comprising wid and subsequence. Results of the sort are given under the post-sorted column in Fig. 2a with wids bolded. The least significant digit resides on the left in our example. Using the LSD variant of radix sort preserves the order of the input strings. For the examples in Figs. 2a & 2b, the result is strings of a cluster are automatically sorted in non-decreasing order of their fitness values. This free sort, provided the input strings are sorted in non-decreasing fitness order, can be advantageous when constructing the V sets.

| | | colspan="6" | All-pairs Hamming Distance | | | | | |
|---|---|---|---|---|---|---|---|---|---|
| sid | | 0 | 1 | 2 | 3 | 4 | 5 | minimal HD | Neighbor set for sid |
| | String | 0101 | 0011 | 1010 | 1101 | 1011 | 1000 | | |
| 0 | 0101 | 0 | 2 | 4 | 1 | 3 | 3 | 1 | {3} |
| 1 | 0011 | | 0 | 2 | 3 | 1 | 3 | 1 | {4} |
| 2 | 1010 | | | 0 | 3 | 1 | 1 | 1 | {4, 5} |
| 3 | 1101 | | | | 0 | 2 | 2 | 2 | {4, 5} |
| 4 | 1011 | | | | | 0 | 2 | 2 | {5} |
| 5 | 1000 | | | | | | 0 | - | { } |
| | | | | | | | | Mean = 7/5 | |

| S | F | sid₂ | wid | Included subsequence y for p=1100 |
|---|---|---|---|---|
| 0101 | 0 | 000 | 0 | 01 |
| | | | 1 | 01 |
| 0011 | 1 | 001 | 0 | 00 |
| | | | 1 | 11 |
| 1010 | 2 | 010 | 0 | 10 |
| | | | 1 | 10 |
| 1101 | 3 | **011** | **0** | **11** |
| | | | 1 | 01 |
| 1011 | 4 | 100 | 0 | 10 |
| | | | 1 | 11 |
| 1000 | 5 | 101 | 0 | 10 |
| | | | 1 | 00 |

| Processed cluster | Working set of sids | sid pair (a, b) | HD(a, b) |
|---|---|---|---|
| A | 1 | - | - |
| B | 1, 0 | (0, 1) | 2 |
| C | 1, 2, 4, 5 | (1, 2) | 2 |
| | | **(1, 4)** | **1** |
| | | (1, 5) | 3 |
| | | **(2, 4)** | **1** |
| | | **(2, 5)** | **1** |
| | | (4, 5) | 2 |
| D | 5, 3 | **(3, 5)** | **2** |
| E | 5 | - | - |
| F | 5, 0, 3 | **(0, 3)** | **1** |
| | | (0, 5) | 3 |
| | | (3, 5) | 2 |
| G | 5, 2 | (2, 5) | 1 |
| H | 5, 1, 4 | (1, 4) | 1 |
| | | (1, 5) | 3 |
| | | (4, 5) | 2 |

| Pre-sorted Table | Post-sorted Table | | sid |
|---|---|---|---|
| 00000001 | 00001**0**00 | A | 1 |
| 00000101 | 00000**0**01 | B | 0 |
| 00001000 | 00010**0**10 | C | 2 |
| 00001111 | 00100**0**10 | | 4 |
| 00010010 | 00101**0**10 | | 5 |
| 00010110 | 00011**0**11 | D | 3 |
| 00**011011** | 00101**1**00 | E | 5 |
| 00011101 | 00000**1**01 | F | 0 |
| 00100010 | 00011**1**01 | | 3 |
| 00100111 | 00010**1**10 | G | 2 |
| 00101010 | 00001**1**11 | H | 1 |
| 00101100 | 00100**1**11 | | 4 |

| sid | b-minimal HD | Vp,w(sid) p = 1100 |
|---|---|---|
| 0 | 1 | {3} |
| 1 | 1 | {4} |
| 2 | 1 | {4, 5} |
| 3 | 2 | {5} |
| 4 | 2 | {5} |
| 5 | - | { } |
| | Mean = 7/5 | |

**Fig. 2a** A *bliss* demonstration. N = 4, |S| = 6, w = N/2, p = 1100, and the sid of a string is also its fitness. Colors are used to highlight similarity in a cluster.

There are eight clusters, marked A to H, in the post-sorted table. Members in a cluster have identical wid and subsequence. The records in cluster C for example, all end with '010'. This implies that the strings in cluster C all have '10' as the first two characters, which indeed '1010', '1011' and '1000' do. In fact, cluster C is the maximal set of strings in S with '10' in the front. Cluster G is a cluster of one since its string '1010' is the only one with '10' in the back.

To construct the V sets for each string, i.e. the set of neighbor strings for s with b-minimal HD to s, we start from the topmost record in the post-sorted table and work all the way down the table, cluster by cluster. Since A is a single member cluster, its string is merged with strings from the next cluster, i.e. cluster B. This produces the working set of sids {0, 1}. For every unique pair (a, b) of sids in a working set where F(b) > F(a), we compute the hamming distance HD(a, b). The objective is to decrease the hamming distance found so far for each string. After processing cluster B, string 1 is left un-neighbored and it is carried over to the next working set which includes the strings from cluster C. The upper right table in Fig. 2a shows all the pairs processed and their respective HD. At the end, we obtain the V sets in the table on the lower right in Fig. 2a. After processing cluster B, $V_{p,w}(0) = \{2\}$, but this is subsequently replaced by {3} when cluster F is processed since string 3 is closer to string 0 than string 2.

| S | F | $sid_2$ | wid | Included subsequence y for p=0110 |
|---|---|---|---|---|
| 0101 | 0 | 000 | 0 | 10 |
|  |  |  | 1 | 01 |
| 0011 | 1 | 001 | 0 | 01 |
|  |  |  | 1 | 01 |
| 1010 | 2 | 010 | 0 | 01 |
|  |  |  | 1 | 10 |
| 1101 | 3 | **011** | **0** | **10** |
|  |  |  | 1 | 11 |
| 1011 | 4 | 100 | 0 | 01 |
|  |  |  | 1 | 11 |
| 1000 | 5 | 101 | 0 | 00 |
|  |  |  | 1 | 10 |

| Processed cluster | Working set of sids | sid pair (a, b) | HD(a, b) |
|---|---|---|---|
| A | 5 | - | - |
| B | 5, 1, 2, 4 | (1, 2) | 2 |
|  |  | (1, 4) | 1 |
|  |  | (1, 5) | 3 |
|  |  | (2, 4) | 1 |
|  |  | (2, 5) | 1 |
|  |  | (4, 5) | 2 |
| C | 5, 0, 3 | (0, 3) | 1 |
|  |  | (0, 5) | 3 |
|  |  | (3, 5) | 2 |
| D | 5, 0, 1 | (0, 1) | 2 |
|  |  | (0, 5) | 3 |
|  |  | (1, 5) | 3 |
| E | 5, 2 | (2, 5) | 1 |
| F | 5, 3, 4 | (3, 4) | 2 |
|  |  | (3, 5) | 2 |
|  |  | (4, 5) | 2 |

| Pre-sorted Table | Post-sorted Table |  | sid |
|---|---|---|---|
| 00000010 | 00101000 | A | 5 |
| 00000101 | 00001001 | B | 1 |
| 00001001 | 00010001 |  | 2 |
| 00001101 | 00100001 |  | 4 |
| 00010001 | 00000010 | C | 0 |
| 00010110 | 00011010 |  | 3 |
| 00**011010** | 00000101 | D | 0 |
| 00011111 | 00001101 |  | 1 |
| 00100001 | 00010110 | E | 2 |
| 00100111 | 00101110 |  | 5 |
| 00101000 | 00011111 | F | 3 |
| 00101110 | 00100111 |  | 4 |

| sid | b-minimal HD | $V_{p,w}(sid)$ p = 0110 |
|---|---|---|
| 0 | 1 | {3} |
| 1 | 1 | {4} |
| 2 | 1 | {4, 5} |
| 3 | 2 | {4, 5} |
| 4 | 2 | {5} |
| 5 | - | { } |
|  | Mean = 7/5 |  |

**Fig. 2b** A *bliss* demonstration. N = 4, |S| = 6, w = N/2, p = 0110, and the sid of a string is also its fitness. Colors are used to highlight similarity in a cluster.

The V sets produced using p=1100 in Fig. 2a do not match perfectly with the ideal neighbor sets set out in table at the top of Fig. 2a. In contrast, the V sets produced using p=0110 in Fig. 2b do. Due to the way the strings are mixed by *bliss* when p = 0110, string 3 gets paired with string 4 but this encounter does not happen when p = 1100. So different patterns can produce different V sets, and this can be used to reduce the error in V sets by running *bliss* with multiple randomly generated patterns and combining the best results for each string. However we may be able to do better than this by observing the identity level of S (Fig. 3).

The identity level of a set of strings $S = \{0,1\}^N$ is a vector whose elements denote the proportion of 1's at each string position. In Fig. 3, the identity level of the leftmost column is 4/6 since 4 out of the 6 strings in S has a '1' at position 0. Pattern p = 0110 (and its complement 1001) keep columns 0 and 3, i.e. the two columns with the highest identity level, together in their subsequences. As a result, it is more likely that highly similar strings will be clustered together, thus increasing their chance of being paired off when constructing the V sets. Most pairings occur between strings of the same cluster. In example Fig. 2b, strings 3 and 4 appear in cluster F and get paired off when *bliss* processes cluster F. Also, *bliss* produced 6 clusters when p = 0110 (Fig. 2b), but 8 clusters when p = 1100 (Fig. 2a).

| String position or column | 0 | 1 | 2 | 3 |
|---|---|---|---|---|
| String 0 (sid=0) | 0 | 1 | 0 | 1 |
| String 1 | 0 | 0 | 1 | 1 |
| String 2 | 1 | 0 | 1 | 0 |
| String 3 | 1 | 1 | 0 | 1 |
| String 4 | 1 | 0 | 1 | 1 |
| String 5 | 1 | 0 | 0 | 0 |
| Number of 1's per column | 4 | 2 | 3 | 4 |
| **Identity level** | **4/6** | **2/6** | **3/6** | **4/6** |
| "good" pattern, w = N/2 | 1 | 0 | 0 | 1 |
|  | 0 | 1 | 1 | 0 |

**Fig. 3** Identity level of S

To complete the investigation of all possible patterns when w=N/2 for the example problem in Fig. 2a, we produce the *bliss* outcome for p = 0101 in Fig. 4. The V sets are identical to those for p = 1100. Hence, p = 0110 (or its complement 1001) do give the best results. However, *bliss* inspects (computes HD for) only 13 pairs when p = 0101, as opposed to 15 when p = 1100. When p = 0110, bliss also inspects 15 pairs but this is not equivalent to an all-pairs comparison as some pairs are repeated. Measures to avoid this repetition may be considered in the future. The speed up provided by *bliss* becomes more evident with larger string sets (Figs. 9 & 10).

An all-pairs comparison can be achieved using a degenerate version of *bliss* where w = N/1, and p = 0000. In this degenerate version, there is only one window, wid = 0, and the subsequence for each window is empty since the pattern p = 0000 says to exclude everything for wid = 0. Since all strings have wid = 0 only and the subsequence for wid = 0 is the empty subsequence (represented in Fig. 5 as '0000'), all strings in S belong to the same cluster. Not surprisingly, this produces V sets which are ideal neighbor sets.

| Pre-sorted Table | Post-sorted Table | | sid |
|---|---|---|---|
| 00000011 | 00010000 | A | 2 |
| 00000100 | 00101000 |   | 5 |
| 00001001 | 00001001 | B | 1 |
| 00001101 | 00100001 |   | 4 |
| 00010000 | 00000011 | C | 0 |
| 00010111 | 00011011 |   | 3 |
| 00011011 | 00000100 | D | 0 |
| 00011110 | 00001101 | E | 1 |
| 00100001 | 00011110 | F | 3 |
| 00100111 | 00101110 |   | 5 |
| 00101000 | 00010111 | G | 2 |
| 00101110 | 00100111 |   | 4 |

| sid | b-minimal HD | $V_{p,w}(sid)$ p = 0101 |
|---|---|---|
| 0 | 1 | {3} |
| 1 | 1 | {4} |
| 2 | 1 | {4, 5} |
| 3 | 2 | {5} |
| 4 | 2 | {5} |
| 5 | - | { } |
|   | Mean = 7/5 |   |

**Fig. 4** *bliss* results for the problem in Fig. 2a with p = 0101.

| Pre-sorted Table | Post-sorted Table | | sid |
|---|---|---|---|
| 00000000 | 000**0**0000 | A | 0 |
| 00100000 | 001**0**0000 |   | 1 |
| 01000000 | 010**0**0000 |   | 2 |
| 01100000 | 011**0**0000 |   | 3 |
| 10000000 | 100**0**0000 |   | 4 |
| 10100000 | 101**0**0000 |   | 5 |

| sid | b-minimal HD | $V_{p,w}(sid)$ p = 0000 |
|---|---|---|
| 0 | 1 | {3} |
| 1 | 1 | {4} |
| 2 | 1 | {4, 5} |
| 3 | 2 | {4, 5} |
| 4 | 2 | {5} |
| 5 | - | { } |
|   | Mean = 7/5 |   |

**Fig. 5** *bliss* results for the problem in Fig. 2a with w = N/4 and p = 0000.

The basic idea behind *bliss* is to compute HD only for those pairs of strings which are highly similar. When w=N/2, strings in a cluster are at least 50% similar. When w = N/1, because of the exclusion rule, a subsequence is 75% of a string, and so strings in a cluster are at least 75% similar. From this, it can seem that better results may be produced by using longer subsequences. However, this is not necessarily the case since a longer subsequence makes it less likely that another string in the set (particularly if the set is a sparse sample) has the identical subsequence for the same window. Consequently, the set of strings may be splintered into many clusters thus restricting the mingling of strings. In the degenerate case when w = N/4 and p = 1111, each string appears exactly once in the table, and forms a cluster by itself. In this case, the input strings themselves get ordered lexicographically from left to right. But this is no reason for producing good V sets, and as Fig. 6 demonstrates, the V sets produced are poor indeed.

On the other hand, larger w values do increase the number of times a sid appears in the table, but its position in the table may not be conducive to produce good V sets. Further, the increased number of records and clusters makes the V sets more expensive to produce. Fig. 7 demonstrates the use of a larger w value on the set of strings S in Fig. 2a. Since w=N/1, the subsequences are now of length 3 and each string appears 4 times in the table. The 24 records are partitioned into 20 clusters, which is more than twice the number of clusters when w = N/2. The resultant V sets are identical to those generated with p = 1100 and with p = 0101, but were more expensive to produce in terms of increased table size and number of pairs evaluated.

| Pre-sorted Table | Post-sorted Table | | sid |
|---|---|---|---|
| 00000101 | 001**00**011 | A | 1 |
| 00100011 | 000**00**101 | B | 0 |
| 01001010 | 101**01**000 | C | 5 |
| 01101101 | 010**01**010 | D | 2 |
| 10001011 | 100**01**011 | E | 4 |
| 10101000 | 011**01**101 | F | 3 |

| sid | b-minimal HD | Vp,w(sid) p = 1111 |
|---|---|---|
| 0 | 2 | {1} |
| 1 | 3 | {5} |
| 2 | 1 | {5} |
| 3 | 2 | {5} |
| 4 | 2 | {5} |
| 5 | - | { } |
| | Mean = 10/5 | |

**Fig. 6** *bliss* results for the problem in Fig. 2a with w = N/4 and p = 1111.

| Pre-sorted Table | Post-sorted Table | | sid |
|---|---|---|---|
| 00000001 | 000**00**001 | A | 0 |
| 00001010 | 001**00**011 | B | 1 |
| 00010011 | 101**00**100 | C | 5 |
| 00011101 | 011**00**101 | D | 3 |
| 00100011 | 010**00**110 | E | 2 |
| 00101001 | 100**00**111 | F | 4 |
| 00110001 | 001**01**001 | G | 1 |
| 00111011 | 000**01**010 | H | 0 |
| 01000110 | 101**01**100 | I | 5 |
| 01001101 | 010**01**101 | J | 2 |
| 01010100 | 100**01**101 | | 4 |
| 01011010 | 011**01**110 | K | 3 |
| 01100101 | 001**10**001 | L | 1 |
| 01101110 | 000**10**011 | M | 0 |
| 01110111 | 010**10**100 | N | 2 |
| 01111101 | 101**10**100 | | 5 |
| 10000111 | 100**10**101 | O | 4 |
| 10001101 | 011**10**111 | P | 3 |
| 10010101 | 101**11**000 | Q | 5 |
| 10011011 | 010**11**010 | R | 2 |
| 10100100 | 001**11**011 | S | 1 |
| 10101100 | 100**11**011 | | 4 |
| 10110100 | 000**11**101 | T | 0 |
| 10111000 | 011**11**101 | | 3 |

| sid | b-minimal HD | Vp,w(sid) p = 3021 |
|---|---|---|
| 0 | 1 | {3} |
| 1 | 1 | {4} |
| 2 | 1 | {4, 5} |
| 3 | 2 | {5} |
| 4 | 2 | {5} |
| 5 | - | { } |
| | Mean = 7/5 | |

**Fig. 7** *bliss* results for the problem in Fig. 2a with w = N/1 and p = 3021.

In an earlier design of *bliss*, we used inclusion instead of exclusion to build the subsequences. For instance, given p = 3021, the second subsequence (wid = 1 or $01_2$) of string '0101' using exclusion is '010' (Fig. 7), but using inclusion is '1' (Fig. 8). Then to assess similarity between a pair of strings, we would count the number of windows or subsequences they have in common and the most similar strings would be those with the highest number of windows in common (top-right table in Fig. 8). Strings habituating the same cluster have at least one subsequence in common. For example, only strings 1, 2, 4 and 5 have '0' at position 1, and so are placed in the same cluster, A. Strings 1 and 4 find themselves sharing a cluster 3 times, once in cluster A, then again in cluster D and finally in cluster F. On inspection, strings 1 and 4, which respectively are '0011' and '1011', have three subsequences in common, namely '0', '1', and '1'. However, we found computing this table too costly for large sets of strings, and ultimately not necessary with the use of exclusion. There may be some differences in the V sets produced using these two different approaches, but the exclusion approach produces *b-walks* satisfactory for our purpose (sections 2 & 3).

| Pre-sorted Table | Post-sorted Table | | sid |
|---|---|---|---|
| 000001 | 001**000** | A | 1 |
| 000011 | 010**000** | | 2 |
| 000100 | 100**000** | | 4 |
| 000110 | 101**000** | | 5 |
| 001000 | 000**001** | B | 0 |
| 001011 | 011**001** | | 3 |
| 001101 | 010**010** | C | 2 |
| 001110 | 101**010** | | 5 |
| 010000 | 000**011** | D | 0 |
| 010010 | 001**011** | | 1 |
| 010101 | 011**011** | | 3 |
| 010111 | 100**011** | | 4 |
| 011001 | 000**100** | E | 0 |
| 011011 | 011**100** | | 3 |
| 011100 | 101**100** | | 5 |
| 011111 | 001**101** | F | 1 |
| 100000 | 010**101** | | 2 |
| 100011 | 100**101** | | 4 |
| 100101 | 000**110** | G | 0 |
| 100111 | 001**110** | | 1 |
| 101000 | 010**111** | H | 2 |
| 101010 | 011**111** | | 3 |
| 101100 | 100**111** | | 4 |
| 101111 | 101**111** | | 5 |

| | Processed cluster | | | | | | | | Most |
|---|---|---|---|---|---|---|---|---|---|
| sid | A | B | C | D | E | F | G | H | similar sid |
| 0 | | 3 | | 1, 3, 4 | 3, 5 | | 1 | | 3 |
| 1 | 2, **4**, 5 | | | 3, **4** | | 2, **4** | | | 4 |
| 2 | 4, 5 | | 5 | | | 4 | | 3, 4, 5 | 4, 5 |
| 3 | | | | 4 | 5 | | | 4, 5 | 4, 5 |
| 4 | 5 | | | | | | | 5 | 5 |
| 5 | | | | | | | | | - |

A fitter string most similar with string 1 is string 4 since string 4 shares a cluster with string 1 more times than any other string.

| sid | b-minimal HD | $V_{p,w}(sid)$ p = 3021 |
|---|---|---|
| 0 | 1 | {3} |
| 1 | 1 | {4} |
| 2 | 1 | {4, 5} |
| 3 | 2 | {4, 5} |
| 4 | 2 | {5} |
| 5 | - | { } |
| | Mean = 7/5 | |

**Fig. 8** Finding similar strings using included subsequence. The set of strings is those in Fig. 2a, w = N/1, and p =3021.

As mentioned and demonstrated before (Figs. 2a & 2b), there may be one or more strings in S which have smaller Hamming distance to s, but they may not be detected by *bliss*. To reduce this error, *bliss* is run multiple times with a different randomly generated p each time, and the best $V_{p,w}(sid)$ found for each sid are combined. Another option is to vary w, the length of an excluded subsequence. However, decreasing w (and thus increasing the length of the included subsequence) increases running time without necessarily decreasing overall b-minimal HD or producing better V sets (Fig. 7). In fact, we found that when $|S| << 2^N$, w = N/2 gives the best results (Fig. 9), provided |S| is large enough. For instance, over the five randomly generated samples of $|S| = 2^{20}$ strings of length N=32, the mean (standard deviation) running time on a 64-bit Linux machine was 134.4 (3.21) seconds when w = 8, and 68.4(1.52) seconds when w = 16. The mean (standard deviation) b-minimal HD per sample was 8.822(0.226) and 4.629(0.003) respectively. Average b-minimal HD per sample is also affected by sample size |S|. When N=64 for example, increasing the sample size from $2^{20}$ to $2^{24}$ did not decrease the mean b-minimal HD per sample (Fig. 9 bottom right). This effect of sample size on b-minimal HD is not unexpected and is not specific to *bliss*.

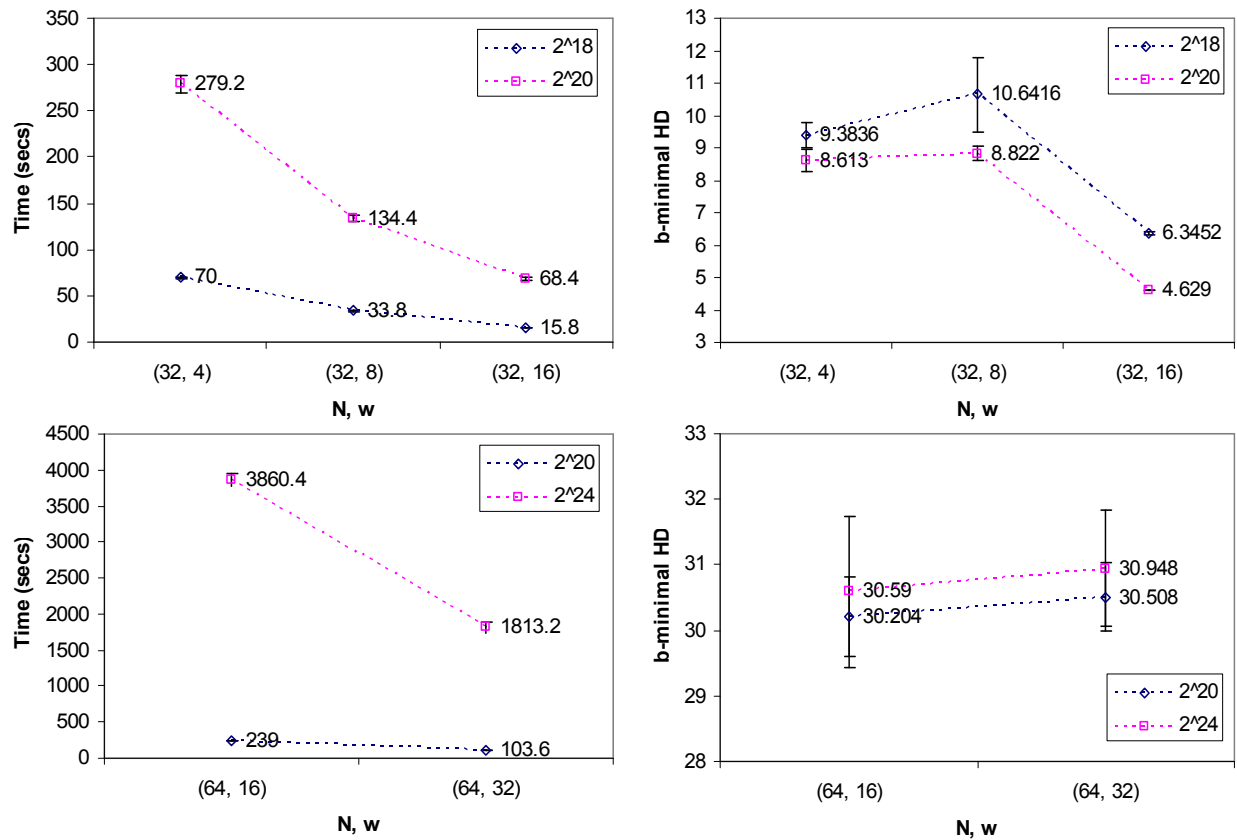

**Fig. 9** *bliss* results averaged over 5 runs for each (N, w) combination. In general, using w=N/2 and larger samples give better results, i.e. takes less time and produces smaller average b-minimal HD. Error bars show one standard deviation from the mean. Programs are compiled with g++ and run on a 64-bit Linux machine with close to 100% CPU allocation.

To demonstrate the speed up effect of *bliss*, Fig. 10 shows time taken to compute Hamming distance between all-pairs of strings in a sample. Each string is stored in a C++ bitset of length 64. It takes 32.275 hours (116,190 seconds) to process a set of randomly generated $2^{20}$ $\{0, 1\}^{32}$ strings.

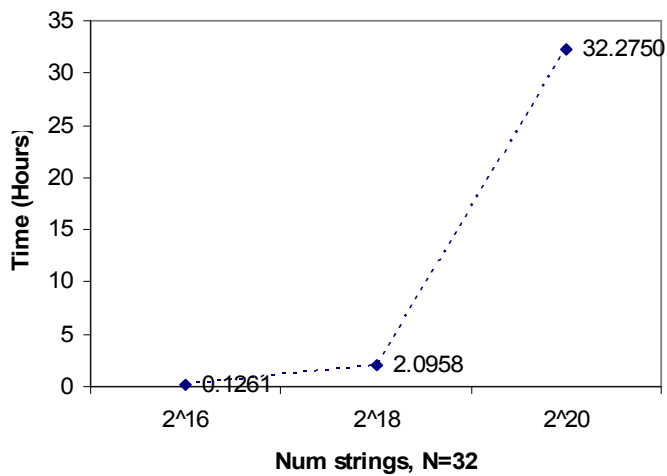

**Fig. 10** Time to (read in and) compute all-pairs Hamming distance by XOR bit implementation. Programs are compiled with g++ and run on a 64-bit Linux machine with close to 100% CPU allocation.

## 2. Testing *b-walks* on enumerated search spaces

The specific purpose of introducing *bliss* is to speed up the construction of slow adaptive walks which are then used to probe a fitness landscape, e.g. to detect search points with local optimum potential (PLOPs). To this end, we ran the *plef* test described in Section 2.2 of ref. [1] using *b-walks*, i.e. slow adaptive walks constructed from *bliss* output V sets. The *plef* test assesses the overlap between the set of actual local optimum points using 1-bit flip neighbourhood **A**, and the set of PLOPs **P** as follows: $|A \cap P| / |A \cup P|$. Ideally, this overlap would be 1.0 for enumerated search spaces. PLOPs are points with local optimum potential by virtue of them having out-going step sizes which are relatively larger then in-coming step sizes (this comparison is done using the *los* formulae given in ref. [1]).

For example, from the V sets in Fig. 2b, the following *b-walks* can be constructed: 0 → 3 → 4 → 5, 1 → 4 → 5, 2 → 5, 3 → 5, and 4 → 5. Like slow adaptive walks, there is a *b-walk* starting from every point in a sample (string in S) except the fittest point(s). The next point in a *b-walk* is chosen at random from the V set of the current point. Every *b-walk* terminates when it reaches a point with an empty V set, which in the example is point (string) 5. Point 3 is a PLOP because its out-going step sizes (3 → 4 and 3 → 5 both have a step size of 2) are larger than its in-coming step sizes (0 → 3 has a step size of 1). Point 3 which is '1101' is at least fitter than one of its 1-bit flip neighbors, i.e. point '0101'. The fitness values of its other 3 neighbors is unknown, however if all '1001', '1111' and '1100' are less or equally fit than '1101', then point 3 has *plef* = 1.0, and is an actual local optimum point in a 1-bit flip neighborhood. The overlap statistic is a measure of how well PLOPs coincide with points having *plef*=1.0 and vice versa.

The objective of conducting the *plef* test is to see if *bliss* produces comparable results to an all-pairs search, in terms of the aforementioned overlap statistic. And indeed, it does. Table 1 reports the overlap statistic averaged over 30 runs for each problem tested. We use the same set of test problems as ref. [1], i.e. NK(16, 4), NK(16, 8), NK(16, 12), HIFFC and HIFFM. We find that the overlap statistics achieved using *b-walks* are close to 1.0. We compared the *bliss* results with previous results in ref. [1] which were obtained by constructing slow adaptive walks from an all-pairs search, which are also close to 1.0.

**Table 1.** The overlap statistic averaged over 30 runs for each problem tested, with standard deviation reported in brackets. $N = 16$, $|S| = 2^{16}$, and $w = 8$.

| Test problem | *bliss* w = 8 | all-pairs |
|---|---|---|
| NK(16, 4) | 0.953963 (0.080272) | 0.97875 (0.017318) |
| NK(16, 8) | 0.981562 (0.005176) | 0.978766 (0.005263) |
| NK(16, 12) | 0.969183 (0.003033) | 0.970011 (0.003203) |
| HIFFC | 0.995313 (0.008359) | 1 (0) |
| HIFFM | 0.998361 (0.005002) | 1 (0) |

Since we now have a faster way to construct slow adaptive walks, it is now more feasible to perform the *plef* test on a larger enumerated search space. We do this with the NK problem for $N = 22$, and K values of 6, 10, 14 and 18. The *b*-walks are constructed from V sets produced using randomly generated patterns for $w = 11$ and $|S| = 2^{22}$. Table 2 reports the overlap statistic averaged over 30 runs for each problem tested. We find that the overlap statistics are close to 1.0, implying that *b-walks* are still meaningful in larger search spaces.

**Table 2.** The overlap statistic averaged over 30 runs for each problem tested, with standard deviation reported in brackets. $N = 22$, $|S| = 2^{22}$, and $w = 11$.

| Test problem | *bliss* w = 11 |
|---|---|
| NK(22, 6) | 0.990866 (0.00229) |
| NK(22, 10) | 0.983473 (0.001137) |
| NK(22, 14) | 0.976311 (0.000773) |
| NK(22, 18) | 0.967733 (0.000684) |

It took less than 2 hours to find the V sets for a problem (Fig. 11) which is a considerable speed up over an all-pairs search (Fig. 10). The $2^{22}$ input strings are fed to *bliss* in lexicographical order and were not sorted by fitness beforehand. Randomly generated patterns where used to create subsequences of length 11. Each string is stored in a C++ bitset of length 64.

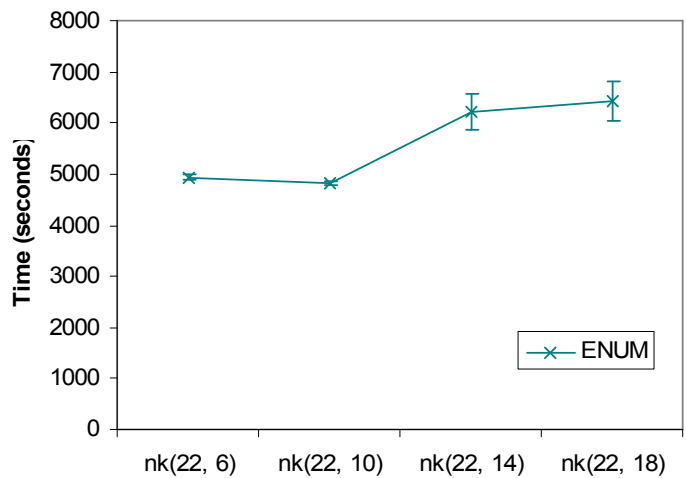

**Fig. 11** Average time (in seconds) for *bliss* to find V sets for $2^{22}$ {0, 1} strings using w = 11. Error bars show 95% confidence level. *bliss* was run 30 times for each of the four NK problems with a randomly generated pattern each time.

## 3. Application of *b-walks* on sampled search spaces

For even larger search spaces, we would like to be able to work on a subset of the search space to grasp the relative ruggedness of one fitness landscape[1] relative to another. To this end, we build on our previous work where we explored the use of Wang-Landau sampling [1] and analyzed the step patterns of slow adaptive walks [3]. We use the NK 22 problems in Table 2. For each problem, 30 instances were created and for each instance an AWL sample of a certain size, and a RAND sample of the same size as the AWL sample, were generated. A RAND sample is produced by selecting points from a search space uniformly at random without replacement. An AWL sample is the set of points visited in one run of the Wang-Landau algorithm. Step statistics produced using AWL samples are compared with those produced from corresponding RAND samples and ENUM samples. An ENUM sample comprises all $2^{22}$ strings. Fig. 12 summarizes our method to produce the step statistics.

The Wang-Landau algorithm [4] does a random walk and accepts moves that visit areas of a search space (defined in terms of fitness bins) which hitherto have been less explored. This gives some level of assurance that all regions of a search space are sampled, and that the sample reflects the fitness distribution or density of states of a search space[2]. For the NK 22 test problems, we ran the Wang-Landau algorithm with the following parameter values: (i) points in the search space are partitioned into fitness bins of width 0.1 all of which could be filled, (ii) the modification factor *f* starts at *e* and is reduced as $f_{t+1} = f_t^{0.5}$ each time the histogram which tracks the number of visits to each fitness bin is flat, (iii) the histogram is considered flat if every fitness bin has been visited at least 0.90 of the number of visits averaged over all bins, (iv) epsilon is $10^{-9}$, and (v) the move operator is 1 bit-flip. The algorithm terminates when the modification factor is less than epsilon, or when the sample size reaches the specified maximum. Three maximum sample size values were used: $2^{16}$, $2^{18}$ and $2^{20}$. We use the exponents to label the AWL samples and their RAND counterparts. E.g. AWL_16 is an AWL sample with a maximum size of $2^{16}$, and RAND_16 is a RAND sample with a maximum size of $2^{16}$.

---

[1] The neighborhood of a point in these fitness landscapes is defined by its nearest fitter set of points or is estimated by its *bliss* V set.

[2] There is an analogy between the problem addressed in this paper, i.e. of trying to understand fitness landscapes via sampling, and the tale of the blind men and the elephant. Wang-Landau sampling provides a way to touch all the different parts of a search space (the elephant), and we believe this reduces the risk of composing an incomplete picture of the search space.

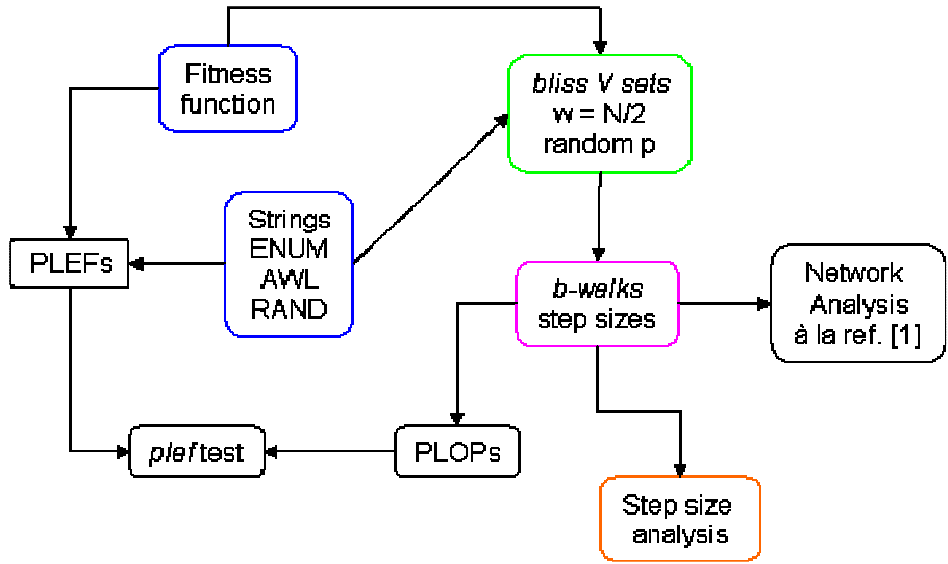

**Fig. 12** Schematic of our method. The step statistics are produced by analyzing the step size taken by slow adaptive walks, or by *b-walks* when *bliss* is used to speed up the construction of slow adaptive walks as is the case here. *b-walks* are constructed from V sets generated by *bliss* which in turn uses a set of strings and their fitness values as input. The *plef* test was described in section 2. It requires as input, PLEFs which are points that are not less fit than all its 1-bit flip neighbors, and PLOPs which are points that have local optima potential.

The step statistics were introduced in [3] to capture fitness landscape difficulty from the view point of a stochastic search algorithm which can modify its move operator and only moves to fitter points. A fitness landscape is deemed more difficult if the search algorithm has to exert more effort to reach a global optimum. For instance, if it needs to adjust its move operator frequently and from a large set of possible moves. The step statistics measure various aspects of the effort expanded by the search algorithm.

In what follows, we define the step statistics and refer to Figs. 13, 14 and 15 which present the step statistics of *b-walks* obtained from the various samples. The graphs in Figs. 13, 14 and 15 serve three purposes. First, they show how the various step statistics change with K and thereby search difficulty. The ENUM plots represent these definitive relationships. Second, they show that AWL plots resemble ENUM plots more closely than RAND plots. And third, they show the effect of sample size. Larger sample sizes produce better results. Both the AWL and RAND plots get closer to the ENUM plots as sample size increases. Nonetheless, the AWL plots are more similar in shape to the ENUM plots than the RAND plots, even when the maximum sample size is $2^{16}$. Hence it is possible to work with a subset of the search space to grasp the relative ruggedness of one fitness landscape relative to another.

Compression ratio *cr1* measures compressibility of walks in terms of steps, and is *cwlen* / *wlen*. Walk length (*wlen*) is the number of steps taken in a walk. Compressed walk length (*cwlen*) is the number of steps in a walk whose steps have been compressed by replacing consecutive steps of the same size with a single step of the size. For example, a walk *w* with steps ⟨1, 1, 2, 3, 2, 2, 2, 5⟩$_w$ is compressed to ⟨1, 2, 3, 2, 5⟩$_{cw}$. Since *wlen* = 8 and *cwlen* = 5, *cr1* = 5/8. Compression ratio *cr2* measures compressibility of walks in terms of distance, and is *cwdist* / *wdist*. Walk distance (*wdist*) is the sum of step sizes taken in a walk. Compressed walk distance (*cwdist*) is the sum of step sizes in a compressed walk. For the previous example, since *wdist* = 18 and *cwdist* = 13, *cr2* = 13/18. Both compression ratios lie in (0.0, 1.0] with 1.0 denoting un-compressible walks. Fig. 13 shows that walks become less compressible, both *cr1* and *cr2* increase, as K increases.

Adaptive length (*adaptlen*) is the length of the longest sequence of same sized steps in a walk. The longest sequence of same sizes steps in the previous example is ⟨2, 2, 2⟩, hence its *adaptlen* is 3. Fig. 14 (left) shows adaptive lengths decreasing as K increases. The suggested inverse relationship between *adaptlen* and both *cr1* and *cr2* agrees with the expectation that more compressible walks have longer adaptive lengths. Less compressible walks and shorter adaptive lengths indicate that more frequent changes to the move operation is necessary to reach a global optimum, thereby demanding more effort from a search algorithm. From a stochastic search algorithm design view point, the compression ratio (either *cr1* or *cr2*) can be seen as the probability of changing the move operator after each move.

*Average step size* is the average step size of all moves made for a problem instance. Fig. 14 (right) shows average step size increases as K increases. Taking larger steps incurs more uncertainty as there are many more reachable but not necessarily fitter neighbors. A search algorithm may spend a lot of time at a search point trying out different combinations. For a distance *d*, there are $\binom{N}{d}$ combinations. This is also a reason to prefer taking smaller steps over larger ones, i.e. slow adaptive walks.

*Step size variation* is the number of unique step sizes taken in a walk. The step size variation of the walk in the previous example is 4 since it makes moves of 1, 2, 3 and 5 step sizes. Fig. 15 (left) shows step size variation increasing with K. *Step size range* is the difference between the maximum and minimum step size taken in a walk. The step size range of the walk in the previous example is 5 − 1 = 4. Fig. 15 (right) shows step size range increasing with K. A more varied step size implies a larger set of move operations that a stochastic search algorithm needs to consider. This coupled with a larger step size range increases uncertainty as to the right move operation to make at a given time.

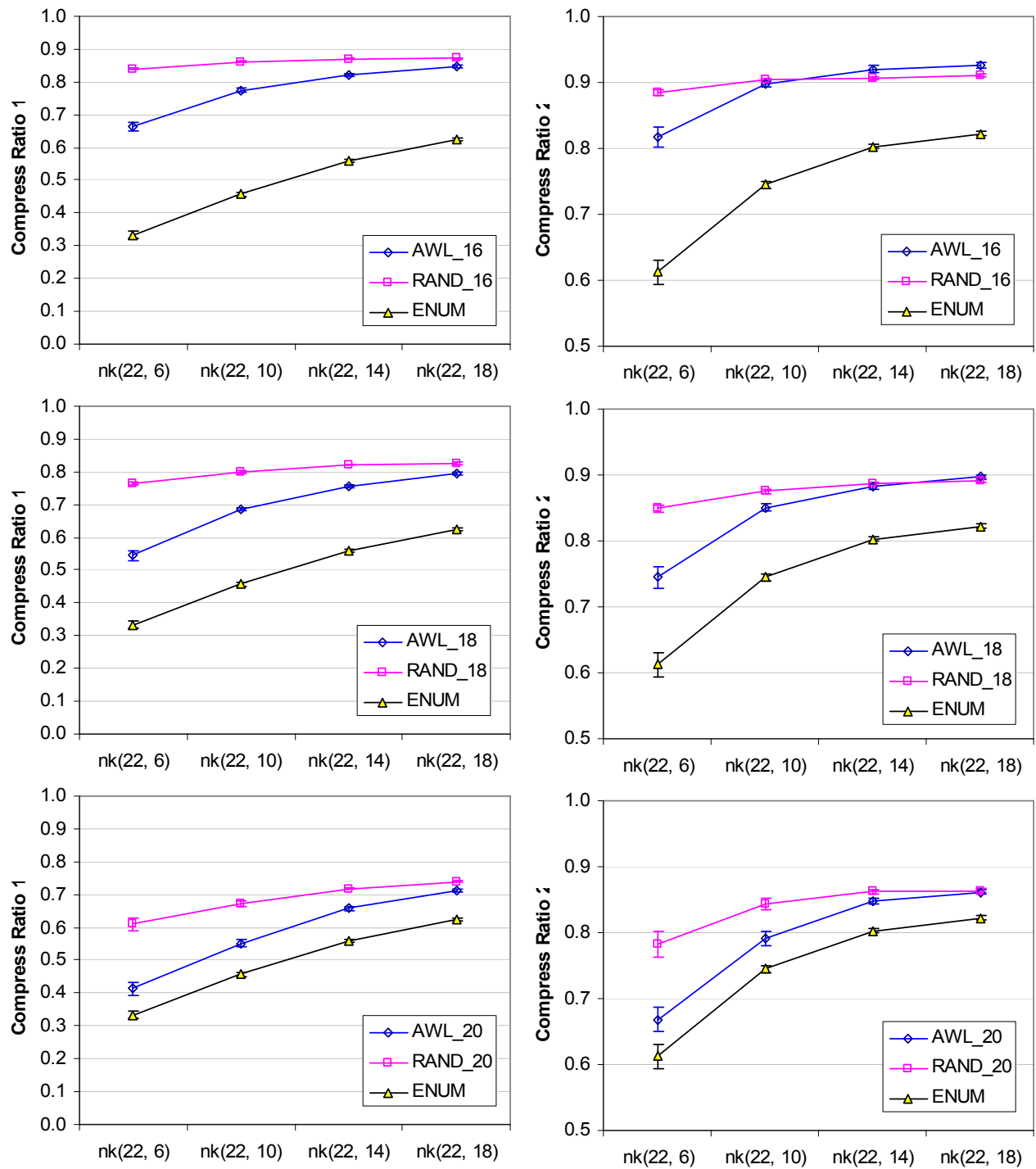

**Fig. 13** Compress ratio 1 (*cr1*) and compress ratio 2 (*cr2*) reflects the compressibility of walks (see text for further explanation). Walks with a ratio close to 1.0 are less compressible. The compressibility of each walk is measured and averaged over all walks per NK instance, and then averaged again over all instances per NK problem. Error bars indicate 95% confidence interval around the final average values.

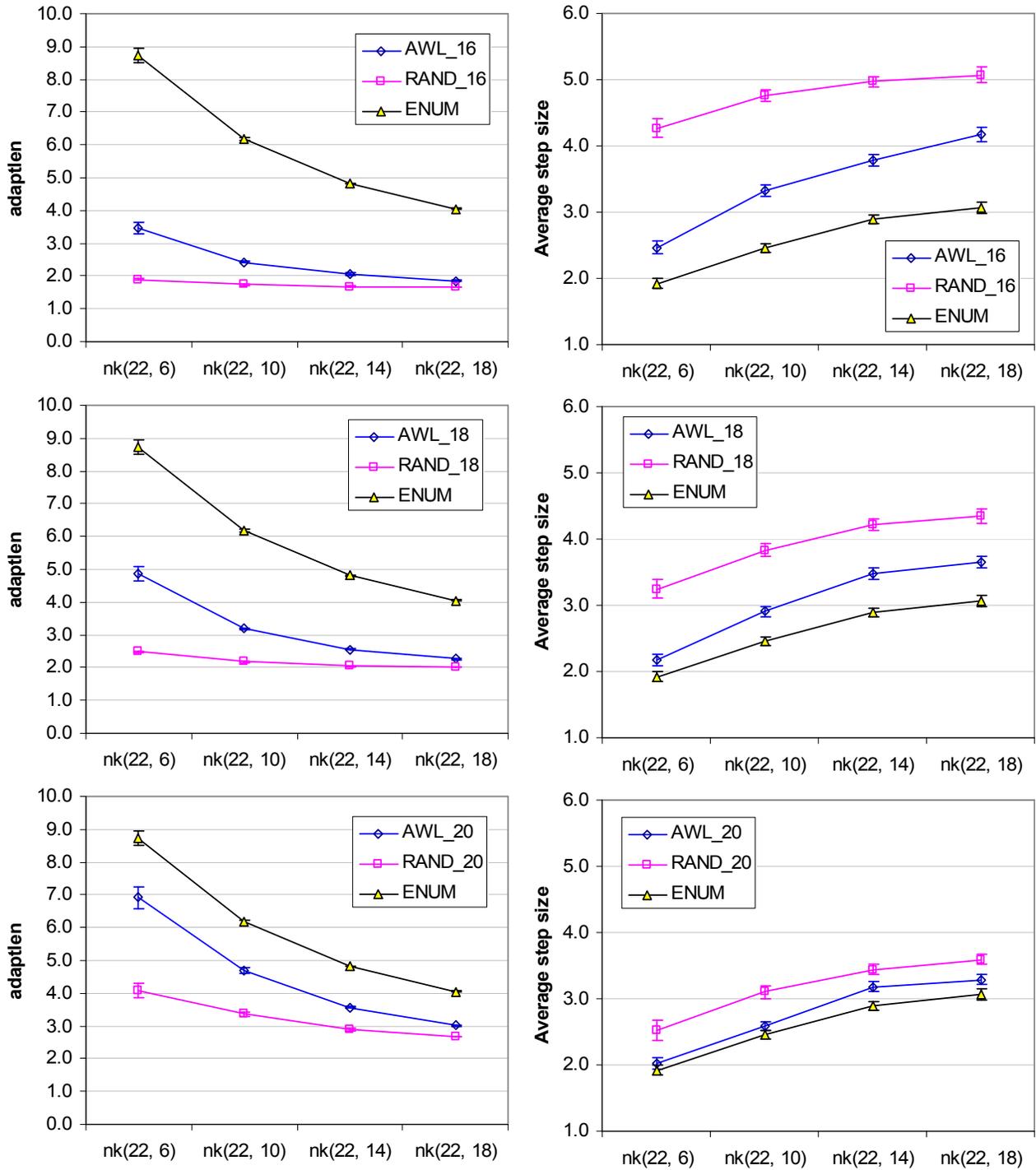

**Fig. 14** adaptlen is the length of the longest sequence of same sized steps in a walk. It is measured for each walk, then summarized over all walks per NK instance, and then summarized again over all instances per NK problem. Average step size is obtained by first taking the average step size of all steps taken for a problem instance, and then averaging these values of over all instances per NK problem. Error bars indicate 95% confidence interval around the final average values.

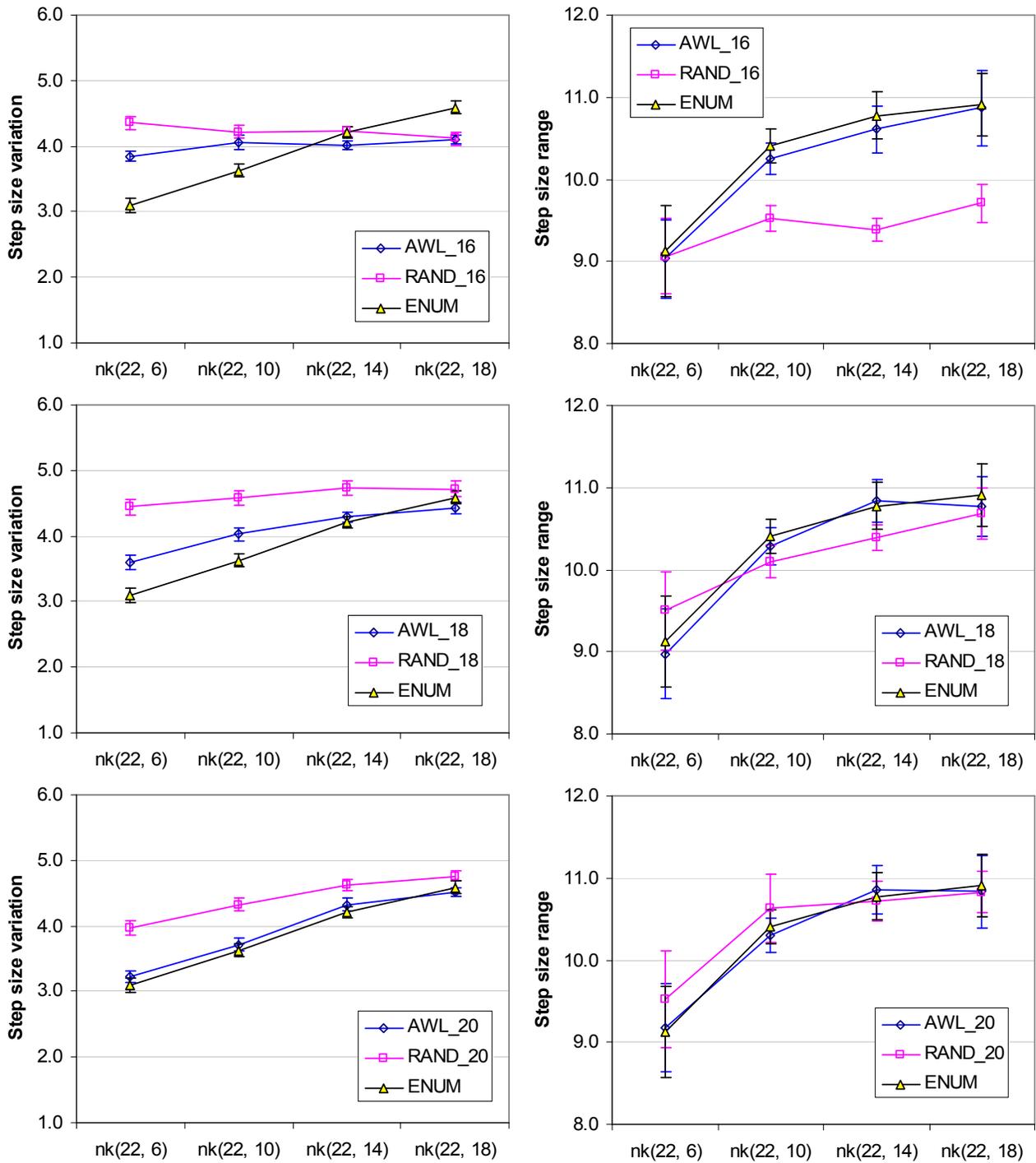

**Fig. 15** Step size variation is the number of unique step sizes taken in a walk. Step size range is the difference between the maximum and minimum step size taken in a walk. These measurements are made for each walk, then summarized over all walks per NK instance, and then summarized again over all NK instances per NK problem. Error bars indicate 95% confidence interval around the final average values.

In short, slow adaptive walks trace out ideal pathways in a fitness landscape. The step statistics describe various aspects of these ideal pathways and assumes for now that all pathways are equally likely.

The step statistics for the NK 22 problems reveal that problems with larger K have less compressible walks, shorter adaptive lengths, larger average step size, more step size variation and larger step size range. All these characteristics increase the necessary choices available to a stochastic search algorithm, i.e. gives it more rope. And thus a search problem becomes more difficult for a stochastic search algorithm.

## 4. Cost of *b-walks* and AWL samples

In this section we discuss the cost of generating slow adaptive walks using *bliss*, and of producing AWL samples for the NK 22 problems studied in section 3. It is not surprising that it took significantly less time to produce V sets for the AWL samples (Fig. 16 left) than for ENUM (Fig. 11). Nonetheless, the time expanded by *bliss* still increases exponentially with problem size (Fig. 16 right), albeit at a much slower rate than the all-pairs version (Fig. 10).

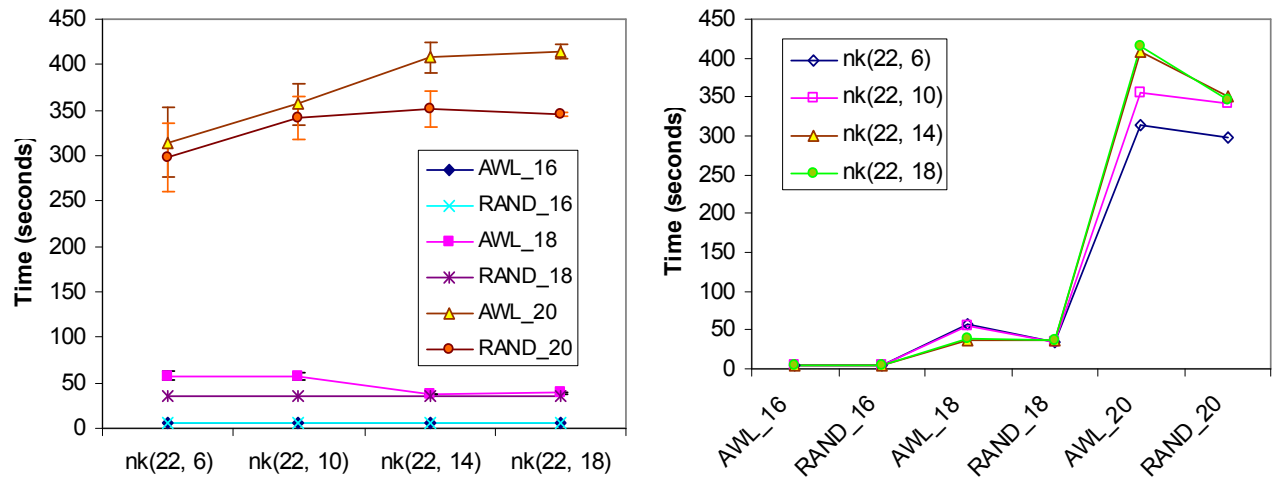

**Fig. 16** Average time (in seconds) for *bliss* to find V sets for AWL and RAND samples using w = 11 and a randomly generated pattern each time. Error bars show 95% confidence level.

There may be many points visited in a Wang-Landau run but not included in an AWL sample. These additional points are due to revisits by the random walker. For the NK 22 problems tested, the number of points visited (Evaluations) is 3 to 5 times the number of points included in an AWL sample (Sample size) (Fig. 17). More damning still is that the number of points visited can exceed the number of points in the entire search space, e.g. AWL_20 for NK (22, 14) and for NK (22, 18). However, this may not be as bad as it seems because the time to generate AWL samples is far less than the time to generate V sets for ENUM samples (Table 3). In the end, it still takes less time to produce step statistics starting with AWL_20 than with ENUM (Table 3) and we saw in section 3 that the AWL_20 step statistics come quite close the ENUM statistics.

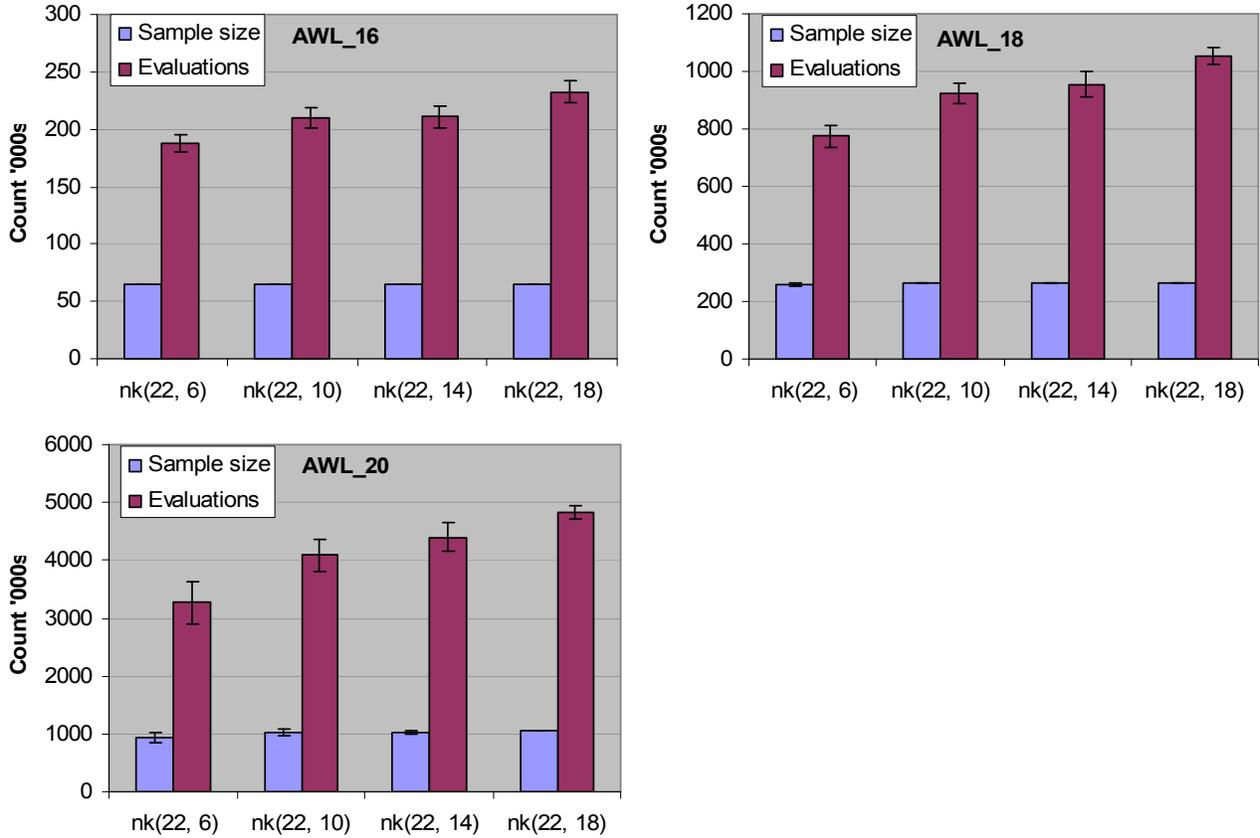

**Fig. 17** Average sample sizes versus the average number of points visited (number of function evaluations made) to generate the samples. Error bars show 95% confidence level.

**Table 3** Time in hours to complete each stage for all 30 instances of a problem, averaged over the four NK problems. One standard deviation around the mean is given in parentheses. Sample generation for ENUM is the time to enumerate all $2^{22}$ binary strings. '0' denotes negligible time. The *bliss* stage is the most costly out of the three stages, and the ENUM samples take the most time to complete all three stages. With fewer points, the AWL samples also have fewer walks to make and to analyze.

| Sample size | Sample generation | *bliss* | Walks and steps |
|---|---|---|---|
| ENUM    $2^{22}$ = 4,194,304 | Assume is 0 | 45.5625 (7.049) | 1.125 (0.25) |
| AWL_16  $2^{16}$ = 65,536 | 0.75 (0) | 0 | 0 |
| AWL_18  $2^{18}$ = 262,144 | 0.75 (0) | 0.375 (0.144) | 0 |
| AWL_20  $2^{20}$ = 1,048,576 | 1.5 (0.204) | 3.0625 (0.427) | 0.5 (0) |

But the high evaluations to sample size ratios raise the issue of fair comparison for the RAND samples. To address this, we generated RAND_16b and RAND_18b samples whose sizes equal the number of points visited by their respective AWL counterparts. The RAND_16b and RAND_18b step statistics are reported in Figs. 18, 19 and 20. With a larger set of points to work with, the RAND_16b and RAND_18b plots come closer to the ENUM plots, but stay parallel to the RAND_16 and RAND_18 plots and so do not take on the shape of the AWL nor the ENUM plots. Thus, the AWL plots are more indicative of the differences in search difficulty between the NK 22 problems, and this sensitivity can be detected with relatively small sample sizes, e.g. AWL_16. To conclude, the combination of Wang-Landau sampling and slow adaptive walks is an effective and economical way to explore a fitness landscape.

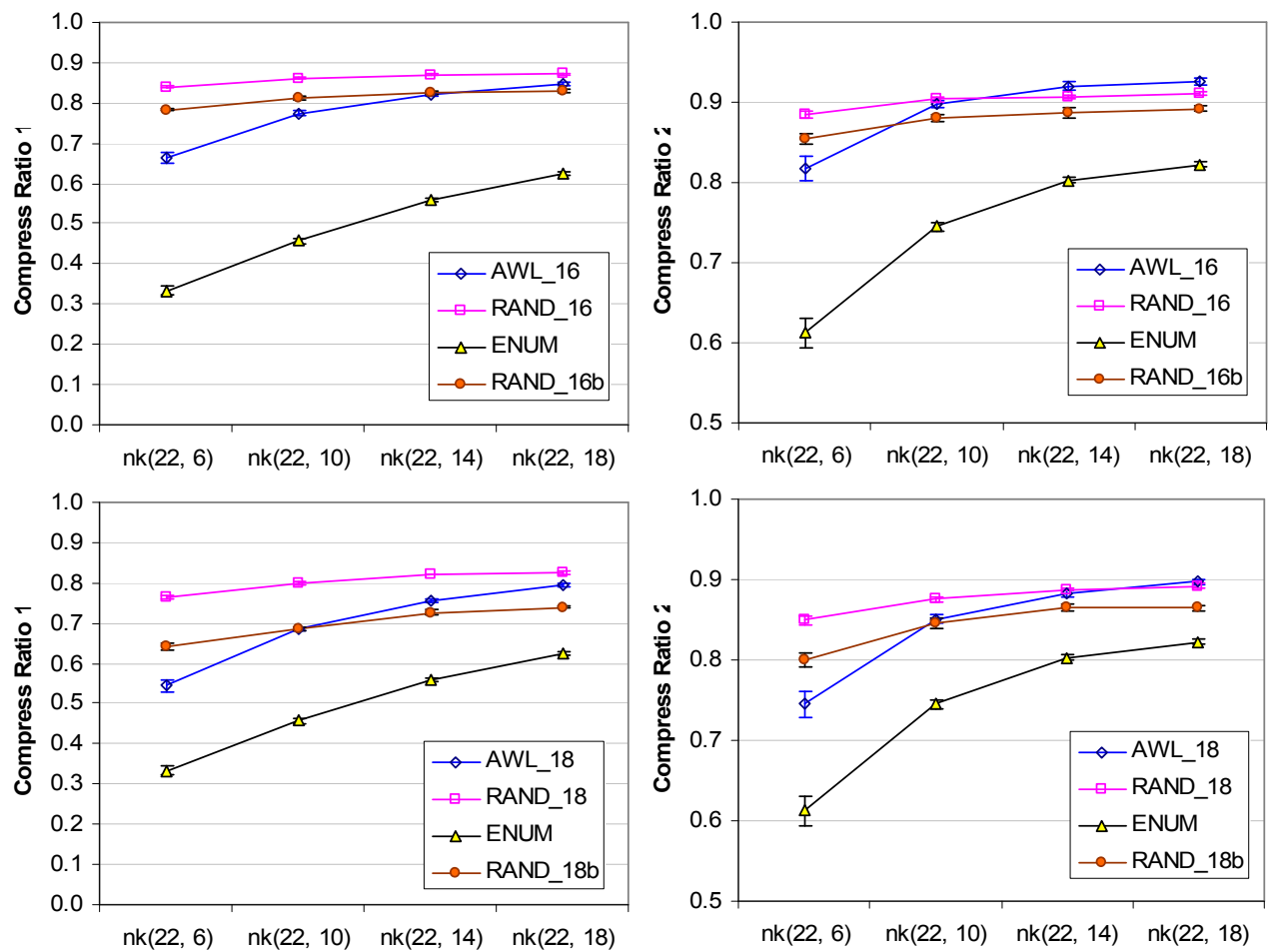

**Fig. 18** Except for the RAND_16b and RAND_18b plots, these graphs are identical to those in Fig. 13.

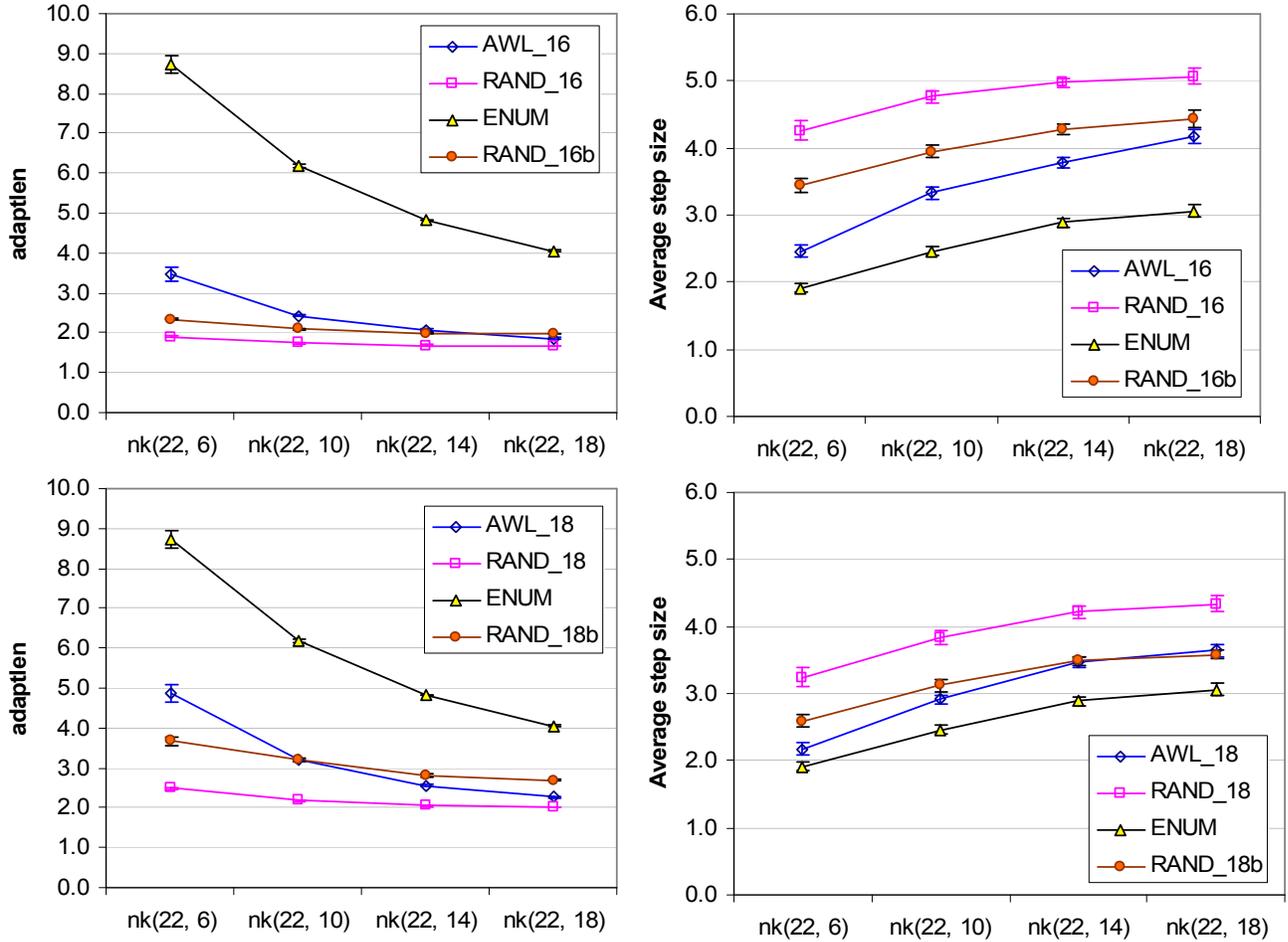

**Fig. 19** Except for the RAND_16b and RAND_18b plots, these graphs are identical to those in Fig. 14.

## 5. Summary

(i) It is possible to work with a subset of the search space to grasp the relative ruggedness of one fitness landscape relative to another. Our method which combines Wang-Landau sampling, *bliss* and step size statistics is an effective and economical way of accomplishing this. A key component to achieving this economy of cost is the *bliss* algorithm which helped to speed up the construction of slow adaptive walks.

(ii) Slow adaptive walks trace out ideal pathways in a fitness landscape. The step statistics describe various aspects of these ideal pathways and assumes for now that all pathways are equally likely. We observed in [3] and in section 3 that NK problems with larger K have less compressible walks, shorter adaptive lengths, larger average step size, more step size variation and larger step size range. All these characteristics increase the necessary choices available to a stochastic search algorithm, i.e. gives it more rope. And thus the search problem becomes more difficult for the stochastic search algorithm.

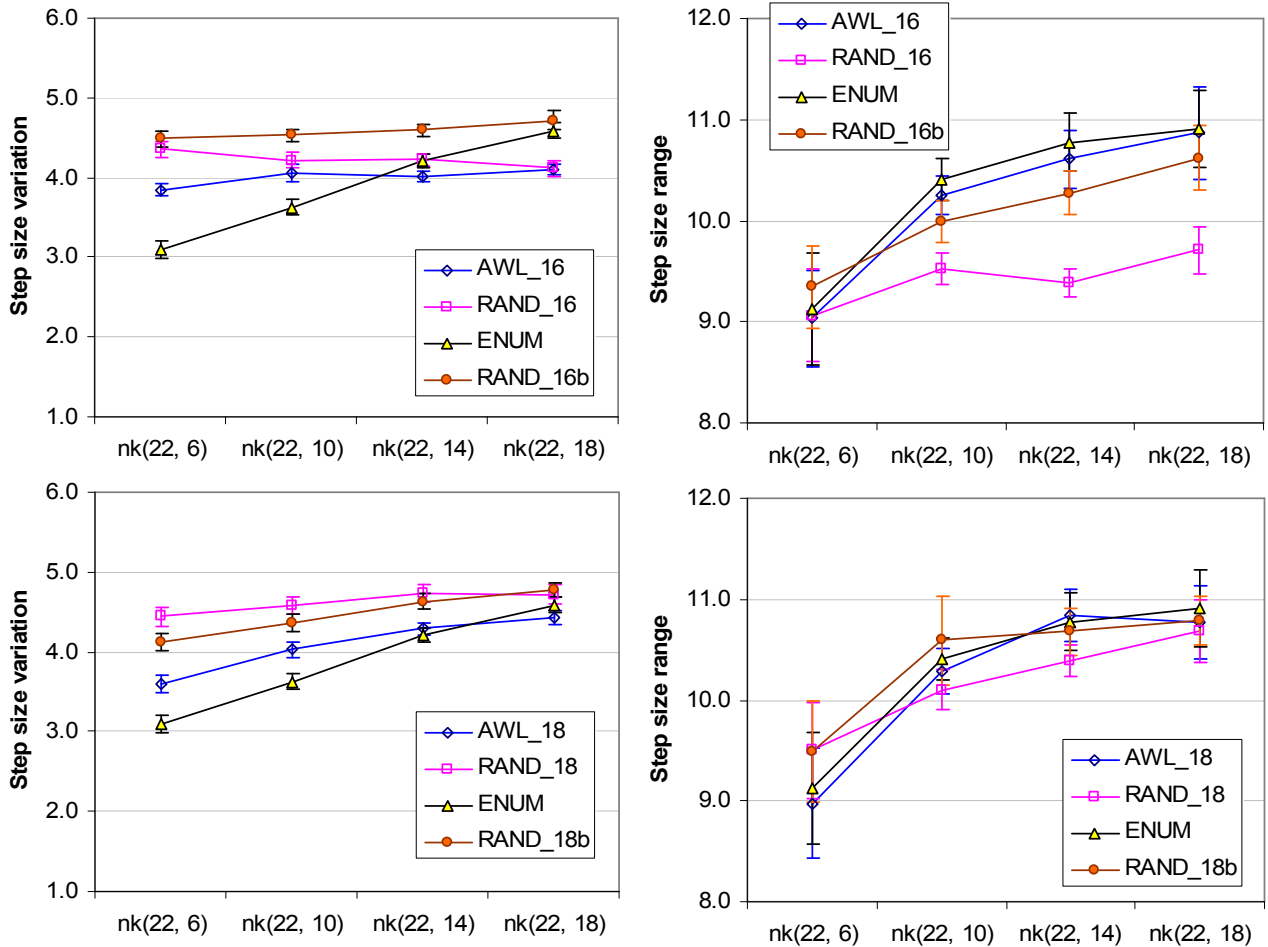

**Fig. 20** Except for the RAND_16b and RAND_18b plots, these graphs are identical to those in Fig. 15.

**Acknowledgements**

My thanks to Dr. P. Grogono for kindly arranging the use of computer resources at Concordia University, Montréal, Québec. Work reported here is independent study although section 4 is the result of responding to some early feedback. This paper is dedicated to the memory of K.S.H.

# Appendix A

Reading point (ii) in the Summary section, we realized that the step statistics could be useful as both description of and prescription for stochastic search algorithms. In section 3 when defining the compression ratios *cr1* and *cr2*, we mentioned that from a stochastic search algorithm design view point, the compression ratio can be seen as the probability of changing the move operator after each move. Here, we sketch how step size statistics might be used to describe the capability of a stochastic search algorithm (we are assuming that a stochastic search algorithm $\mathcal{A}$ is ideally suited for a problem $\mathcal{P}$, and so in essence there is no difference between describing $\mathcal{A}$ and describing $\mathcal{P}$). Within such a scheme, a more capable stochastic search algorithm would have a longer description than a less capable one.

In Table A1, we have the step statistics produced from AWL_18 samples. These are the values plotted in Figs. 13, 14 and 15 for AWL_18. We find that the sum of these four step statistics increases with K, and at this point we might be satisfied to use the sum values to rank capability. A more satisfying alternative is to convert the step statistics into discrete values in some range (there is an opportunity here to emphasize or deemphasize certain step statistics), and create a string from the discreet values which describe the capability of a stochastic search algorithm. More capable stochastic search algorithms would have longer strings to describe them. This idea is illustrated in Table A2. The descriptive strings have special symbols in them to delineate the step statistics, and to guide the alignment of strings. Gaps in the aligned strings are filled with a gap symbol ('0' in Table A2). The length of a description is inversely related to the number of gap symbols in the describing string. A two algorithm example: C5S2V4R9 is a description of the search algorithm for NK (22, 6). When compared with C7S3V4R10, it needs 4 gap symbols (7-5 + 3-2 + 10-9), while C7S3V4R10 needs none. Therefore C5S2V4R9 has a shorter description and is less capable than C7S3V4R10.

**Table A1** AWL_18 **s**tep statistics for the NK 22 problems in section 3.

| K | *cr1* | Avg. step size | Step size variation | Step size range | sum |
|---|---|---|---|---|---|
| 6 | 0.5450 | 2.1703 | 3.5983 | 8.9763 | 15.2900 |
| 10 | 0.6863 | 2.9090 | 4.0293 | 10.2887 | 17.9133 |
| 14 | 0.7570 | 3.4697 | 4.2870 | 10.8423 | 19.3560 |
| 18 | 0.7943 | 3.6497 | 4.4307 | 10.773 | 19.6477 |

**Table A2** Comparing string descriptions of stochastic search algorithms.

| K | C | S | V | R | Aligned description | Num zeroes |
|---|---|---|---|---|---|---|
| 6 | 5 | 2 | 4 | 9 | C11111000A1100V1111R11111111100 | 7 |
| 10 | 7 | 3 | 4 | 10 | C11111110A1110V1111R11111111110 | 3 |
| 14 | 8 | 3 | 4 | 11 | C11111111A1110V1111R11111111111 | 1 |
| 18 | 8 | 4 | 4 | 11 | C11111111A1111V1111R11111111111 | 0 |